\documentclass[letterpaper]{article} 
\usepackage{aaai25}  
\usepackage{times}  
\usepackage{helvet}  
\usepackage{courier}  
\usepackage[hyphens]{url}  
\usepackage{graphicx} 
\usepackage{natbib}  
\usepackage{caption} 
\usepackage{algorithm}
\usepackage{algorithmic}

\usepackage{algorithm}
\usepackage{algorithmic}
\usepackage{subcaption}
\usepackage{amssymb}
\usepackage{amsmath}
\usepackage{xcolor}
\usepackage{colortbl}
\usepackage{booktabs}
\usepackage{multirow} 
\usepackage{microtype}

\usepackage{newfloat}
\usepackage{listings}
\DeclareCaptionStyle{ruled}{labelfont=normalfont,labelsep=colon,strut=off} 
\floatstyle{ruled}
\newfloat{listing}{tb}{lst}{}
\floatname{listing}{Listing}
\title{FlowMamba: Learning Point Cloud Scene Flow with Global Motion Propagation}
\author{
    Min Lin\textsuperscript{\rm 1}, Gangwei Xu\textsuperscript{\rm 2}, Yun Wang\textsuperscript{\rm 2}, Xianqi Wang\textsuperscript{\rm 1}, Xin Yang\textsuperscript{\rm 2,3}\thanks{Corresponding author} \\
}
\affiliations{
    \textsuperscript{\rm 1}School of Artificial Intelligence and Automation, Huazhong University of Science \& Technology\\
    \textsuperscript{\rm 2}School of EIC, Huazhong University of Science \& Technology\\
    \textsuperscript{\rm 3}Hubei Key Laboratory of Smart Internet Technology, Huazhong University of Science \& Technology\\

    \{minlin, gwxu, wangyun, xianqiw, xinyang2014\}@hust.edu.cn
    
}

\usepackage{bibentry}

\begin{document}

\maketitle

\begin{abstract}
Scene flow methods based on deep learning have achieved impressive performance. However, current top-performing methods still struggle with ill-posed regions, such as extensive flat regions or occlusions, due to insufficient local evidence. In this paper, we propose a novel global-aware scene flow estimation network with global motion propagation, named FlowMamba. The core idea of FlowMamba is a novel Iterative Unit based on the State Space Model (ISU), which first propagates global motion patterns and then adaptively integrates the global motion information with previously hidden states. As the irregular nature of point clouds limits the performance of ISU in global motion propagation, we propose a feature-induced ordering strategy (FIO). The FIO leverages semantic-related and motion-related features to order points into a sequence characterized by spatial continuity. 
Extensive experiments demonstrate the effectiveness of FlowMamba, with $21.9\%$ and $20.5\%$ EPE3D reduction from the best published results on FlyingThings3D and KITTI datasets. Specifically, our FlowMamba is the first method to achieve millimeter-level prediction accuracy in FlyingThings3D and KITTI. Furthermore, the proposed ISU can be seamlessly embedded into existing iterative networks as a plug-and-play module, improving their estimation accuracy significantly.
\end{abstract}

%

\section{Introduction}
\label{sec:intro}

Scene flow estimation is the task of calculating three-dimensional motion fields from consecutive frames with various downstream applications, e.g., autonomous driving~\cite{teng2023motion,xu2022attention,xu2023accurate}, robotic manipulation~\cite{seita2023toolflownet}, augmented reality~\cite{wang2023pointshopar}, etc. Nowadays, scene flow estimation still faces many challenges such as in flat regions, slender structures, or occlusions. These areas can be broadly defined as ill-posed areas: \textit{regions with insufficient or even missing geometric features}. Such regions pose apparent challenges in predicting scene flow, as they introduce considerable local ambiguity and unreliable point correspondences between frames.

\begin{figure}[t]
  \centering
  \begin{subfigure}{1.0\linewidth}
        \begin{subfigure}{0.49\linewidth}
            \centering
            \includegraphics[width=1.0\linewidth]{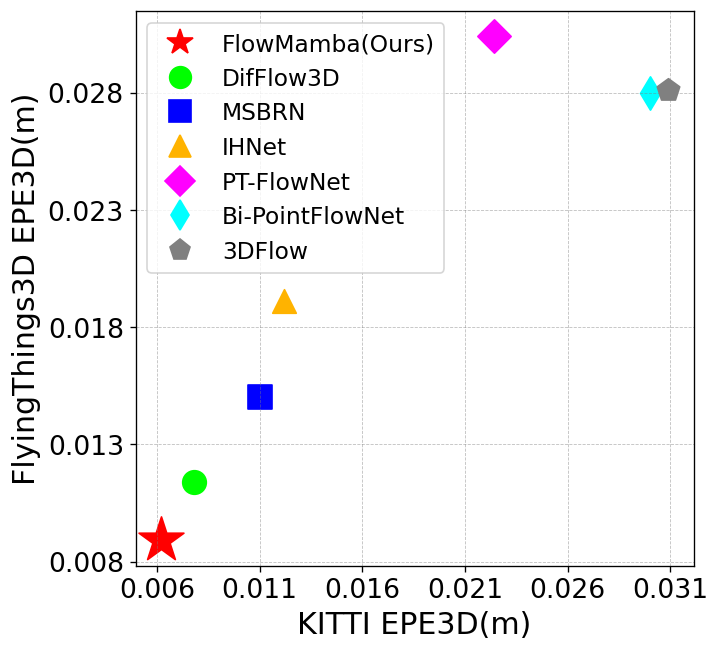}
        \end{subfigure}
        \begin{subfigure}{0.48\linewidth}
            \centering
            \includegraphics[width=1.0\linewidth]{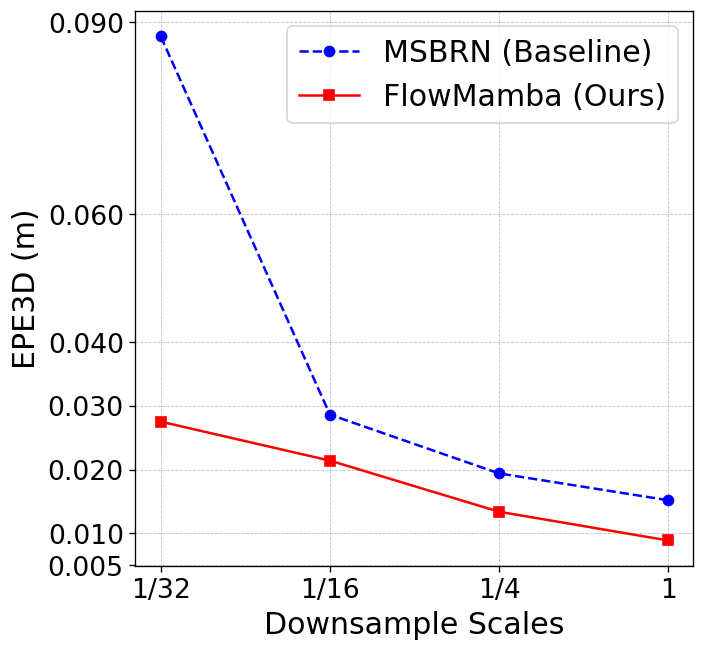}
        \end{subfigure}
        \vspace{1px}
    \end{subfigure}

   \caption{\textbf{Left}: Comparison with state-of-the-art scene flow methods~\cite{liu2024difflow3d,cheng2023multi,wang2023ihnet,fu2023pt,cheng2022bi,wang2022matters} on FlyingThings3D and KITTI. Notably, we achieved millimeter-level precision on both datasets for the first time. \textbf{Right}: Comparison with the accuracy of each layer output on FlyingThings3D. Our FlowMamba can achieve superior results from the coarsest level. In practical applications, adjusting the number of levels and iterations allows for a trade-off between efficiency and accuracy.}
   \label{fig:compare}
   \vspace{-10pt}
\end{figure}

Earlier approaches address the challenge areas by employing neighboring points to recover motion. This can be achieved by using CNNs to learn the relationships between neighboring points~\cite{wu2020pointpwc,wang2022matters} or by constraining the scene flow to exhibit rigid motion within a local region~\cite{li2021hcrf,wang2023ihnet}. However, both methods are limited by the local information available within a small operational window, focusing primarily on local evidence. While local evidence becomes inadequate for recovering hidden motion, it severely reduces the performance of current state-of-the-art methods. Recent studies, such as the work by Lu et al.~\cite{lu2023gma3d}, have investigated the use of non-local approaches to model long-range dependencies between local descriptors, aiming to address the issue of insufficient local evidence. While these approaches help to some extent, they still tend to fail because a severe lack of geometric structure significantly diminishes the representational power of local descriptors.

A potential solution to tackle local ambiguity is to utilize global interpretations, such as learning global relationships through Transformers~\cite{zhang2024gmsf}. However, the quadratic complexity of the attention mechanism imposes significant computational demands and hinders inference speed. Motivated by recent studies \cite{liu2024point,zhang2024point,liang2024pointmamba} that apply structured state space models (SSM) and Mamba to point cloud processing, which presents an effective global receptive field with linear complexity, we introduce a novel scene flow estimation method named FlowMamba. FlowMamba benefits from a novel iterative SSM-based update module (ISU) and a feature-induced ordering (FIO) strategy to efficiently capture long-range motion and model complex patterns, effectively addressing the issues in local ambiguous regions, as illustrated in Figure 4.

Specifically, we claim that the ISU module plays the key role of propagating global information, which motivates two critical designs. Firstly, we advocate for bidirectional sequence modeling to ensure comprehensive information aggregation from all other points. Secondly, the ISU integrates global hidden information in an iterative update process, enhancing the model's capability to capture complex motion patterns. However, the inherent irregularity of point clouds can restrict the propagation of global motion when directly applying the ISU module without ordering, as this irregularity may introduce erroneous spatial relationship during the sequence modeling process. To overcome this limitation, we propose the feature-induced ordering (FIO) strategy. The core concept is to enable the network to construct causal dependencies among points at a higher level, implicitly preserving spatial consistency. Specifically, we harness semantic-related and motion-related features—contextual features, motion information, and the updated hidden information—as crucial cues to generate a score for each point and determine the sequence order.

 We demonstrate that long-range connections facilitated by the ISU in FlowMamba significantly enhance scene flow estimation, especially for addressing the motion of ill-posed regions where local information is inadequate. On the FlyingThings3D dataset~\cite{mayer2016large}, our FlowMamba reaches the state-of-the-art EPE3D with $21.9\%$ and $20.9\%$ lower errors under the non-occluded and occluded scenarios, respectively. And on the real-world KITTI dataset~\cite{geiger2013vision}, FlowMamba improves the generality by decreasing the errors $20.5\%$ and $9.6\%$ on non-occluded and occluded scenarios, respectively. 

 Overall, the contributions of our paper are as follows:
\begin{enumerate}
\item We propose a novel scene flow estimation architecture, named FlowMamba, which incorporates global motion propagation to significantly enhance the robustness of motion estimation.

\item  We introduce a novel iterative SSM-based update
module (ISU), enabling the efficient integration of global motion information in the point cloud. 

\item  We introduce a feature-induced ordering strategy (FIO) to alleviate the impact of point cloud irregularity on global motion propagation.

\item  Our method outperforms existing published methods on FlyingThings3D and KITTI datasets. Especially, our FlowMamba achieves millimeter-level precision on both datasets for the first time. We also verify the universality of our ISU on several scene flow methods.

\end{enumerate}

\section{Related work}
\label{sec:related}

\subsection{Point Cloud-based Scene Flow Estimation.}

Currently, the emergence of deep learning on point cloud ~\cite{qi2017pointnet,qi2017pointnet++,wu2019pointconv,wu2023pointconvformer} has sparked interest in obtaining scene flow directly from point cloud data. FlowNet3D~\cite{liu2019flownet3d} was the pioneering method to address scene flow estimation in point clouds using learning-based techniques. PointPWC-Net~\cite{wu2020pointpwc} employs a coarse-to-fine approach for estimating scene flow in point clouds.
Bi-PointFlowNet~\cite{cheng2022bi} presents a novel architecture for estimating scene flow, which utilizes bidirectional flow~\cite{xu2024hdrflow} embedding layers to enhance flow estimation performance. RMS-FlowNet~\cite{battrawy2022rms} introduces the Patch to Dilated Patch flow embedding design, allowing for more robust scene flow prediction in conjunction with Random-Sampling. 3DFlow~\cite{wang2022matters} introduces an all-to-all flow embedding layer with backward reliability verification based on the coarse-to-fine construct. IHNet~\cite{wang2023ihnet} proposes a resampling scheme to alleviate the problem of poor correspondence. Nevertheless, these methods face limitations, such as error accumulation in early steps and a tendency to miss fast-moving objects. 

\begin{figure*}[t]
    \centering
    \includegraphics[width=1.0\linewidth]{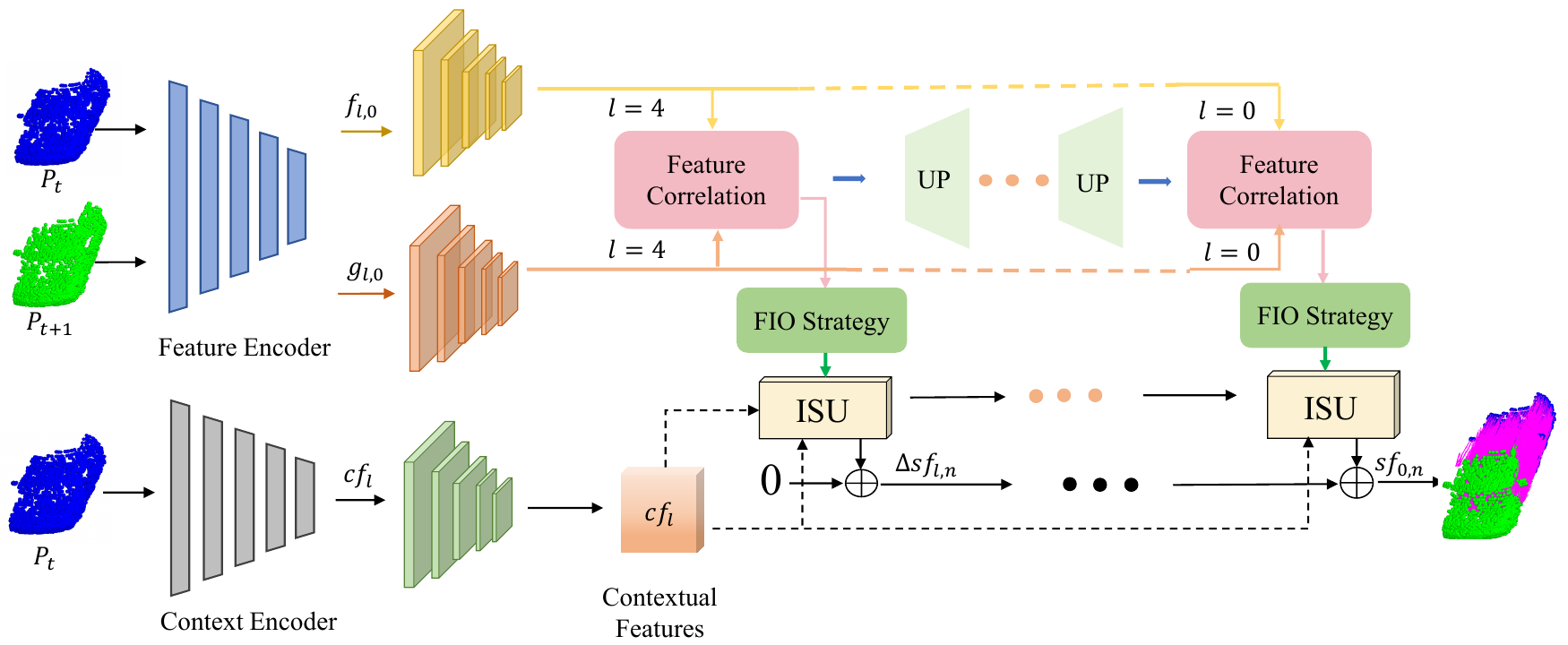}
    \caption{Overview of our proposed FlowMamba. Feature encoders abstract the point clouds to obtain the multi-scale point feature and context feature. The correlation features can be obtained by using local cost volumes retrieved from the feature pyramid. The iterative SSM-based update module (ISU) is designed to update the hidden information and scene flow by capturing long-range dependencies and comprehensive motion patterns with global motion propagation. The feature-induced ordering (FIO) strategy is designed to construct reasonable causal dependencies in point cloud.}
    \label{fig:overview network}
    \vspace{-10pt}
\end{figure*}

In recent years, many iterative methods~\cite{kittenplon2021flowstep3d,wei2021pv,gu2022rcp,fu2023pt,xu2023iterative,xu2024igev++,wang2024selective} have gradually become the mainstream of research. PV-RAFT~\cite{wei2021pv} proposes point-voxel correlation fields to handle large and small displacements. RCP~\cite{gu2022rcp} employs a two-stage recurrent network, where 3D flows are optimized at a point-wise level in the first stage and then globally regularized in a recurrent network in the second stage. PT-FlowNet~\cite{fu2023pt} applies transformer \cite{zhao2021point,zhang2020feature} in all functional stages of the task and achieves outstanding results. MSBRN ~\cite{cheng2023multi} proposes an effective and efficient architecture by iterative optimizing coarse-to-fine scene flow. DifFlow3D~\cite{liu2024difflow3d} introduces an uncertainty-aware scene flow estimation network with the diffusion probabilistic model to enhance resilience to challenging cases. \textit{However, how to efficiently enhance matching capabilities in complex regions through global motion propagation remains a challenging issue.}

\subsection{State Space Model in Point Cloud.}
Inspired by continuous state space models in control systems, recently, There has been a significantly increasing focus on the state space models (SSMs)~\cite{gu2021efficiently,gu2023mamba}, which have been proven to have competitive long-range dependency modeling ability. In particular, S4 ~\cite{gu2021efficiently} introduces a normalization technique for the parameters to achieve stable diagonalization, significantly reducing computational overhead and memory consumption. Mamba ~\cite{gu2023mamba} presents a selection mechanism and hardware-aware algorithms, resulting in superior outcomes compared to transformers. Currently, many methods apply mamba to 3D point cloud classification, segmentation tasks, static point cloud generation, and completion~\cite{liu2024point,zhang2024point,liang2024pointmamba,li20243dmambacomplete,mo2024efficient}. PointMamba~\cite{liang2024pointmamba} is the first to apply mamba to point cloud learning. 3DMambaComplete~\cite{li20243dmambacomplete} proposes to use mamba's global modeling capabilities to enhance the understanding of point cloud reconstruction. DiM-3D~\cite{mo2024efficient} uses mamba to improve the scalability of models and the performance of outputting high-resolution voxels. \textit{However, it remains unclear whether the state space model can model dynamic point clouds, i.e., scene flow.}

\section{Methods}
\label{sec:methods}

We based our network design on the successful MSBRN \cite{cheng2023multi} architecture. The overview of our proposed method is shown in Figure \ref{fig:overview network}. For the given two consecutive point clouds $P_t = \{p_i \in \mathbb{R}^3\}_{i=1}^{N_1}$, $P_{t+1} = \{q_j \in \mathbb{R}^3\}_{j=1}^{N_2}$, we estimate the 3D motion vector for each point as ${SF} = \{{sf} \in \mathbb{R}^3\}_{i=1}^{N_1}$.

\begin{figure*}[t]
  \centering
  \includegraphics[width=1.0\linewidth]{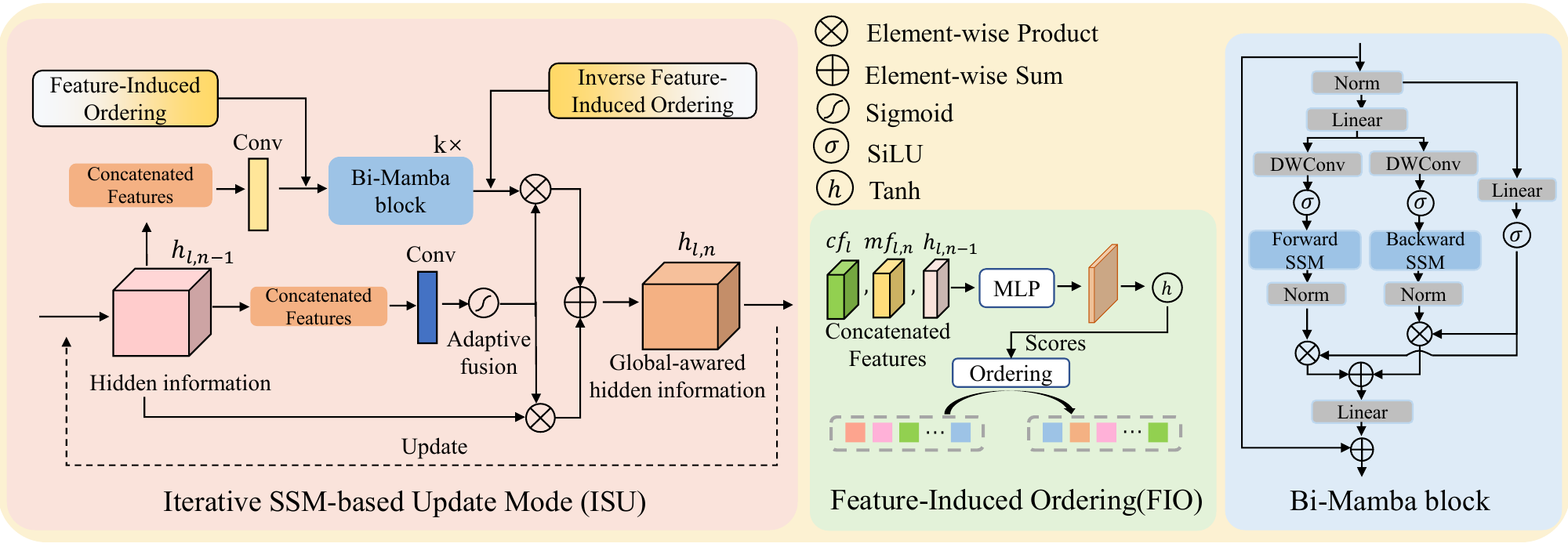}  

   \caption{The architecture of proposed module. Left: Iterative SSM-based Update (ISU) Module. Middle: Feature-Induced Ordering(FIO) strategy. Right: Bi-directional Mamba block (Bi-Mamba).}
   \label{fig:ISU}
   \vspace{-10pt}
\end{figure*}

\subsection{Preliminaries} 
The SSM-based models, such as structured state space sequence models (S4)~\cite{gu2021efficiently} and Mamba~\cite{gu2023mamba}, are motivated by a continuous-time learnable framework for mapping between continuous-time scalar inputs $x(t) \in \mathbb{R}$ and outputs $y(t) \in \mathbb{R}$ through an implicit latent state $h(t) \in \mathbb{R}^N$. 

\begin{equation}
\label{equ:ssm1}
\begin{aligned}
    h'(t) &= \mathbf{A}h(t) + \mathbf{B}x(t), \\
    y(t) &= \mathbf{C}h(t)
\end{aligned}
\end{equation}

Where the $\mathbf{A}$ is the evolution parameter and $\mathbf{B}$, $\mathbf{C}$ is the projection parameters.

The S4~\cite{gu2021efficiently} and Mamba~\cite{gu2023mamba} are the discrete versions of the continuous system. Zero-order hold (ZOH)  discretizes the continuous-time SSM to a discrete-time SSM by using a timescale parameter $\boldsymbol{\Delta}$.

\begin{equation}
\begin{aligned}
    \mathbf{A} &= \exp(\boldsymbol{\Delta} \mathbf{A}), \\
    \mathbf{B} &= (\boldsymbol{\Delta} \mathbf{A})^{-1} (\exp(\boldsymbol{\Delta} \mathbf{A}) - \mathbf{I}) \cdot \boldsymbol{\Delta} \mathbf{B}
\end{aligned}
\end{equation}

In this case the formula (\ref{equ:ssm1}) can be rewritten as:

\begin{equation}
\begin{aligned}
     \mathbf{h}_t = \overline{\mathbf{A}} \mathbf{h}_{t-1} + \overline{\mathbf{B}} \mathbf{x}_t, \quad \mathbf{y}_t = {\mathbf{C}} \mathbf{h}_t
\end{aligned}
\end{equation}

Finally, the models compute output through a global convolution.

\begin{equation}
\begin{aligned}
     \mathbf{K} = (\mathbf{C}\overline{\mathbf{B}}, \mathbf{C}\overline{\mathbf{A}}\overline{\mathbf{B}}, \ldots, \mathbf{C}\overline{\mathbf{A}}^{N-1}\overline{\mathbf{B}}), \quad y = x \ast \mathbf{K}
\end{aligned}
\end{equation}

\subsection{Feature Extraction}
Following the design introduced in ~\cite{qi2017pointnet++,wu2019pointconv,cheng2023multi}, we utilize multi-scale point feature extractions as the backbone. Feature encoder and context encoder produces $L$-level pyramid of point features $f_{l,0} \in \mathbb{R}^{N_1 \times C}$, $g_{l,0} \in \mathbb{R}^{N_2 \times C}$, and $cf \in \mathbb{R}^{N_1 \times C}$, starting with the input point clouds at the top. At each level $l$, the farthest point sampling (FPS) subsamples dense points and features to create a sparse set. For each sparse point, k-nearest neighbors (KNN) form a local region. A Pointconv layer~\cite{wu2019pointconv} then dynamically weights and aggregates features from these neighbors to obtain a local feature for each sparse point.

\subsection{Iterative SSM-based Update Module}

The local receptive field in the classical GRU structure hinders the propagation of matching information and the modeling of long-range dependency motions. To tackle this limitation, we propose an iterative update module based on SSM with global motion propagation (ISU).

As illustrated in Figure \ref{fig:ISU},  
for a certain layer $l$, our iterative module takes the coordinates $p_i \in \mathbb{R}^3$ and features $f_{l,0} \in \mathbb{R}^{N_1 \times C_1}$ of the source point cloud, the coordinates $q_j \in \mathbb{R}^3$ and features $g_{l,0} \in \mathbb{R}^{N_2 \times C_1}$ of the target point cloud, context information $cf_{l} \in \mathbb{R}^{N_1 \times C_1}$, motion features $mf_{l,0} \in \mathbb{R}^{N_1 \times C_2}$, flow $sf_{l,0} \in \mathbb{R}^{N_1 \times 3}$, and the hidden information from last layer $h_{l+1,0}$ as input. After $N$ iterations, the updated point cloud features $f_{l,n}$, $g_{l,n}$, flow $sf_{l,n}$, and hidden state $h_{l,n}$ are used as inputs for subsequent scales. 

Our ISU module mainly consists of two parts: global hidden information optimization and adaptive fusion update.

\noindent\textbf{Global hidden information optimization.} 

We stack several layers of bidirectional Mamba (Bi-Mamba) blocks to achieve global information aggregation. Specifically, we first sort the point cloud in each iteration, as detailed in the following section. The sorted hidden information is then fed into bidirectional Mamba blocks for global optimization. Each bidirectional Mamba block utilizes layer normalization (LN), bidirectional Selective SSM, depth-wise convolution (DW)~\cite{chollet2017xception}, and residual connections. The bidirectional Mamba layer is shown on the right of Figure \ref{fig:ISU}, and the output can be present as follows:

\begin{equation*}
\begin{aligned}
    h_{{l}, {n-1}}' &=  \text{LN}( \text{Conv}_{1d}([cf_{l}, mf_{l,n}, h_{l,n-1}])), \\
    h_{{l},n}' &=  \sigma(\text{DW}(\text{Linear}({h_{{l}, {n-1}}'})) \\
    h_{{l},n}'' &=  \sigma((\text{Linear}({h_{{l}, {n-1}}'})) \\
     h_{{l},n} &=  \text{Linear}(\text{Bi-SSM}(h_{{l},n}') \odot {h_{{l},n}''}) + h_{{l},{n-1}}
\end{aligned}
\end{equation*}

Where $h_{{l},n} \in \mathbb{R}^{{N_1} \times C}$ is the output of the $l$-th block in $n$-th iterations, and $\sigma$ indicates $SiLU$ activation~\cite{hendrycks2016gaussian}. 
The bidirectional SSM ~\cite{zhu2024vision,liu2024point} is the key to the Bi-Mamba block.

\noindent\textbf{Adaptive fusion update.} 

The globally optimized hidden information is then fused with the historical information from the previous iteration. Specifically, we use a gated mechanism to adaptively fuse them, resulting in the final global-aware hidden information. The adaptive fusion method is defined as follows:
\begin{equation}
\begin{aligned}
    w_{{l},n} &=  {sigmoid}(\text{Conv}_{1d}([cf_{l}, mf_{l,n}, h_{l,n-1}])) \\
    \hat{h}_{{l},n} &=  (1 - w_{{l},n}) \odot h_{{l},n-1} + w_{{l},n} \odot h_{{l},n}
\end{aligned}
\end{equation}

Where $w_{{l},n}$ denotes the attentive weight, $\odot$ denotes the Hadamard product, $h_{{l},n-1}$ denotes the hidden information from the previous iteration.

\setlength{\tabcolsep}{0.2mm}
\begin{table*}[!t]
        \footnotesize
	\begin{center}
	
		\resizebox{1.00\textwidth}{!}
            {
		\begin{tabular}{lcccccc|cccccc}
		    \hline
			\toprule
			
			&\multicolumn{6}{c|}{FT3D$_{s}$ }&\multicolumn{6}{c}{KITTI$_{s}$ }\\
			 Method & EPE3D\textcolor{red}{$\downarrow$} & Acc3DS\textcolor{green!60!gray}{$\uparrow$} & Acc3DR\textcolor{green!60!gray}{$\uparrow$} & Outliers\textcolor{red}{$\downarrow$}& EPE2D\textcolor{red}{$\downarrow$}& Acc2D\textcolor{green!60!gray}{$\uparrow$}& EPE3D\textcolor{red}{$\downarrow$} & Acc3DS\textcolor{green!60!gray}{$\uparrow$}& Acc3DR\textcolor{green!60!gray}{$\uparrow$}& Outliers\textcolor{red}{$\downarrow$}& EPE2D\textcolor{red}{$\downarrow$}& Acc2D\textcolor{green!60!gray}{$\uparrow$}\\ \midrule

			FlowNet3D   & 0.1136 & 0.4125 & 0.7706& 0.6016& 5.9740 & 0.5692 & 0.1767& 0.3738 & 0.6677  & 0.5271  &  7.2141  & 0.5093  \\
			
			HPLFlowNet      & 0.0804 & 0.6144 & 0.8555& 0.4287& 4.6723 & 0.6764 & 0.1169 & 0.4783  & 0.7776  & 0.4103  &  4.8055  &  0.5938\\
			
			PointPWC-Net    & 0.0588   & 0.7379  & 0.9276 & 0.3424 & 3.2390 &  0.7994 & 0.0694 &  0.7281  &  0.8884 & 0.2648 &3.0062 &0.7673\\
			  
			 HALFlow     &  0.0492  &  0.7850&  0.9468 & 0.3083 &2.7555 &  0.8111 &  0.0622 & 0.7649 &0.9026 & 0.2492 & 2.5140 &  0.8128\\
			FLOT    & 0.0520  &  0.7320 & 0.9270 &  0.3570 & ---&  --- & 0.0560 &  0.7550  &  0.9080  & 0.2420   & ---& ---\\

			PV-RAFT   & 0.0461 & 0.8169  & 0.9574  &  0.2924 & ---& ---  & 0.0560 & 0.8226  & 0.9372  &  0.2163 & ---& ---    \\

            RCP  & 0.0403 & 0.8567 & 0.9635   &   0.1976   &  --- &   --- & 0.0481 & 0.8491 & 0.9448   &   0.1228   &  --- &   ---
            \\
			
			 3DFlow     & 0.0281 & 0.9290 & 0.9817   &   0.1458   &  1.5229 &   0.9279 & 0.0309 &  0.9047  &0.9580    &    0.1612   &    1.1285 &       0.9451
            \\ 
            
             Bi-PointFlowNet     & 0.0280 & 0.9180 & 0.9780   &   0.1430   &  1.5820 &   0.9290 & 0.0300 & 0.9200 & 0.9600   &   0.1410   &  1.0560 &   0.9490
            \\
            
             PT-FlowNet    & 0.0304 & 0.9142 & 0.9814   &   0.1735   &  1.6150 &   0.9312 & 0.0224 & 0.9551 & 0.9838   &   0.1186   &  0.9893 &   0.9667
            \\
             IHNet    & 0.0191 & 0.9601 & 0.9865   &   0.0715   &  1.0918 &   0.9563 & 0.0122 & 0.9779 & 0.9892   &   0.0913   &  0.4993 &   0.9862
            \\

             MSBRN    & 0.0150 & 0.9730 & 0.9920   &   0.0560   &  0.8330 &   0.9700 & 0.0110 & 0.9710 & 0.9890   &   0.0850   &  0.4430 &   0.9850 
            \\
            
            DifFlow3D   & 0.0114 & 0.9836 & 0.9949   &   0.0350   &  0.6220 &   0.9824 & 0.0078 & 0.9817 & 0.9924   &   0.0795   &  0.2987 &   0.9932 
            \\

            Ours  
				&\bf0.0089 & \bf 0.9861
				&\bf 0.9955 & \bf 0.0243	
				& \bf 0.4946 & \bf 0.9848
				& \bf0.0062& \bf0.9876
				&\bf 0.9938 &  \bf0.0741
				&\bf 0.2604 &  \bf 0.9919\\
            
            \bottomrule
            \hline
			
		\end{tabular}
		}
	\vspace{-8pt}
	\caption{\textbf{Comparison results on FlyingThings3D \cite{mayer2016large} and KITTI~\cite{geiger2013vision} datasets without occlusion \cite{gu2019hplflownet}.}  
 Our method has $\bf{21.9}\%$ and $\bf{20.5\%}$ EPE3D reduction respectively compared with previous works. It is worth noting that, for the first time, we achieved millimeter-level precision on both datasets (0.0089 on FlyingThings3D and 0.0062 on KITTI). The best results are in \bf{bold}.}
	\label{table:nocc}
	\end{center}
	\vspace{-6mm}
	
\end{table*}

\subsection{Feature-induced Ordering Strategy}
\label{feature ordering}

The ordering strategy is critical for the effective utilization of SSM. To mitigate the impact of point cloud irregularity on the propagation of global information, We propose a novel feature-induced ordering strategy and sort all points in each iteration. In detail, we generate a score for each point by using three features related to motion estimation: contextual information, correlation feature, and the updated hidden information. We then sort these scores to determine the neighborhoods of the points. 
These features represent the spatial and motion states of the point cloud in the current iteration, revealing the intrinsic relationships between the points. Thus, the FIO strategy ensures the construction of causal dependencies and spatial continuity in the point cloud, thereby facilitating the propagation of features.

\begin{equation}
\begin{aligned}
    {Score} = {tanh}{(\text{MLP}[cf_{l}, mf_{l,n}, h_{l,n-1}])}
\end{aligned}
\end{equation}

Where $cf_{l}$ denotes the context feature, $mf_{l,n}$ denotes the motion features in $n$-th iterations, and $h_{l,n-1}$ denotes the updated hidden information from the ${n-1}$ iterations.

\subsection{The Architecture of FlowMamba}

 We adopt a coarse-to-fine structure to estimate the residual flow at different scales. For each layer, the sparse scene flow $sf_{l+1,n}$, hidden state $h_{l+1,n}$, and $cf_{l+1}$ in the coarse layer are upsampled to dense. Following the approach in \cite{cheng2023multi,liu2024difflow3d}, we then employ a warp operation to iteratively align the source point cloud with the target point cloud using the updated flow field. Subsequently, we implement a bidirectional feature enhancement strategy based on the coordinates and features of the two point clouds. Using SetConv (which includes shared multilayer perceptrons and max-pooling layers), we effectively aggregate features from the other frame to enhance the feature representation of the current point cloud. Finally, the enhanced features are cached to provide a foundation for subsequent iteration processes or scaling.

\subsection{Loss Function}

\noindent\textbf{Flow Supervision.} We follow most of the supervised scene flow approaches to design the loss function~\cite{wang2022matters,cheng2023multi,gu2019hplflownet,gu2022rcp,fu2023pt,wang2021hierarchical}. Specifically, we use the L2 norm between the ground truth flow and the estimated flow in each iteration. 

\begin{equation}
\begin{split}
    L =  \sum_{l=1}^L \alpha^{(l)} \sum_{n=1}^{N^l}\left\|sf_{l,n}^{gt} - sf_{l,n}\right\|_2
\end{split}
\end{equation}
Where $sf_{l,n}$ is the scene flow of the $n^{th}$ iterative estimation. $N$ is the total number of iterations, and $\alpha^{(l)}$ is the weight of the $l^{th}$ layer. We adopt the hyper-parameter setting in~\cite{cheng2023multi,liu2023difflow3d} with $\alpha^{(0)} = 0.16$, $\alpha^{(1)} = 0.08$, $\alpha^{(2)} = 0.04$, $\alpha^{(3)} = 0.02$.

\section{Experiments}
\label{sec:exper}

\subsection{Datasets and Implementation Details}

\noindent\textbf{Datasets.} For a fair comparison, we trained FlowMamba following the previous methods~\cite{wei2021pv,fu2023pt,cheng2023multi} on FlyingThings3D~\cite{mayer2016large} and tested on both FlyingThings3D and KITTI~\cite{geiger2013vision}. FlyingThings3D is a large synthetic dataset, including 19,640 pairs of labeled training samples and 3,824 samples in the test set. We directly evaluated the model on KITTI~\cite{geiger2013vision} without any fine-tuning to validate the generalization ability on the real-world KITTI dataset, which contains 200 pairs of test data.

\setlength{\tabcolsep}{0.2mm}
\begin{table}[t]
	\begin{center}
					
		\resizebox{1.00\columnwidth}{!}
            {
		\begin{tabular}{clccccc}
		    \hline
			\toprule
			
		Dataset& Method   & EPE3D\textcolor{red}{$\downarrow$} & Acc3DS\textcolor{green!60!gray}{$\uparrow$}& Acc3DR\textcolor{green!60!gray}{$\uparrow$}& Outliers\textcolor{red}{$\downarrow$}\\ \midrule

			& FlowNet3D    & 0.169 & 0.254 & 0.579 & 0.789\\			
			
			&  FLOT       & 0.156  &  0.343 & 0.643  & 0.700 \\			
					
			&  FESTA      & 0.111  &  0.431 & 0.744  & --- \\
			
		    \multirow{-1}{*}{\begin{tabular}[c]{@{}c@{}}FT3D$_{o}$ \end{tabular}}

			&  3DFlow    & 0.063 & 0.791 & 0.909  & 0.279 \\ 

            &  Bi-PointFlowNet   & 0.073 & 0.791 & 0.896  & 0.274 \\
            &  MSBRN     & 0.053 & 0.836 & 0.926  & 0.231  \\
            &  DifFlow3D    & 0.043 & 0.891 & 0.944  & 0.133  \\

            & Ours     & \bf0.034 &\bf0.9165 & \bf0.9531 & \bf0.0904 \\ 
			\midrule
			
			& FlowNet3D   & 0.173 & 0.276 & 0.609  & 0.649\\
			
			&  FLOT    & 0.110  &  0.419 & 0.721   & 0.486\\

			&  FESTA     & 0.097  &  0.449 & 0.833  & --- \\

			& 3DFlow       &0.073  &0.819  & 0.890 & 0.261  \\ 

            \multirow{-3}{*}{\begin{tabular}[c]{@{}c@{}}KITTI$_{o}$\end{tabular}}
            &  Bi-PointFlowNet    & 0.065 & 0.769 & 0.906  & 0.264 \\
                       
        &  MSBRN     & 0.044 & 0.873 & 0.950  & 0.208  \\
        
         &  DifFlow3D     & 0.031 & 0.955 & 0.966  & 0.108  \\
   
            & Ours     &\bf0.028  &\bf0.959  & \bf0.971 & \bf0.096  \\ 
			\bottomrule
			\hline
		\end{tabular}
		}
	\vspace{-8pt}
	\caption{\textbf{Comparison results on FlyingThings3D and KITTI datasets with occlusion \cite{liu2019flownet3d}.} Our method outperforms previous works by \textbf{$\bf{20.9}\%$} and \textbf{$\bf{9.6}\%$} in terms of EPE3D. The best results are in \bf{bold}.}
	\label{table:occ}
	\end{center}
	\vspace{-6mm}
\end{table}

\begin{figure*}[t]
\centering
\includegraphics[width=0.99\linewidth]{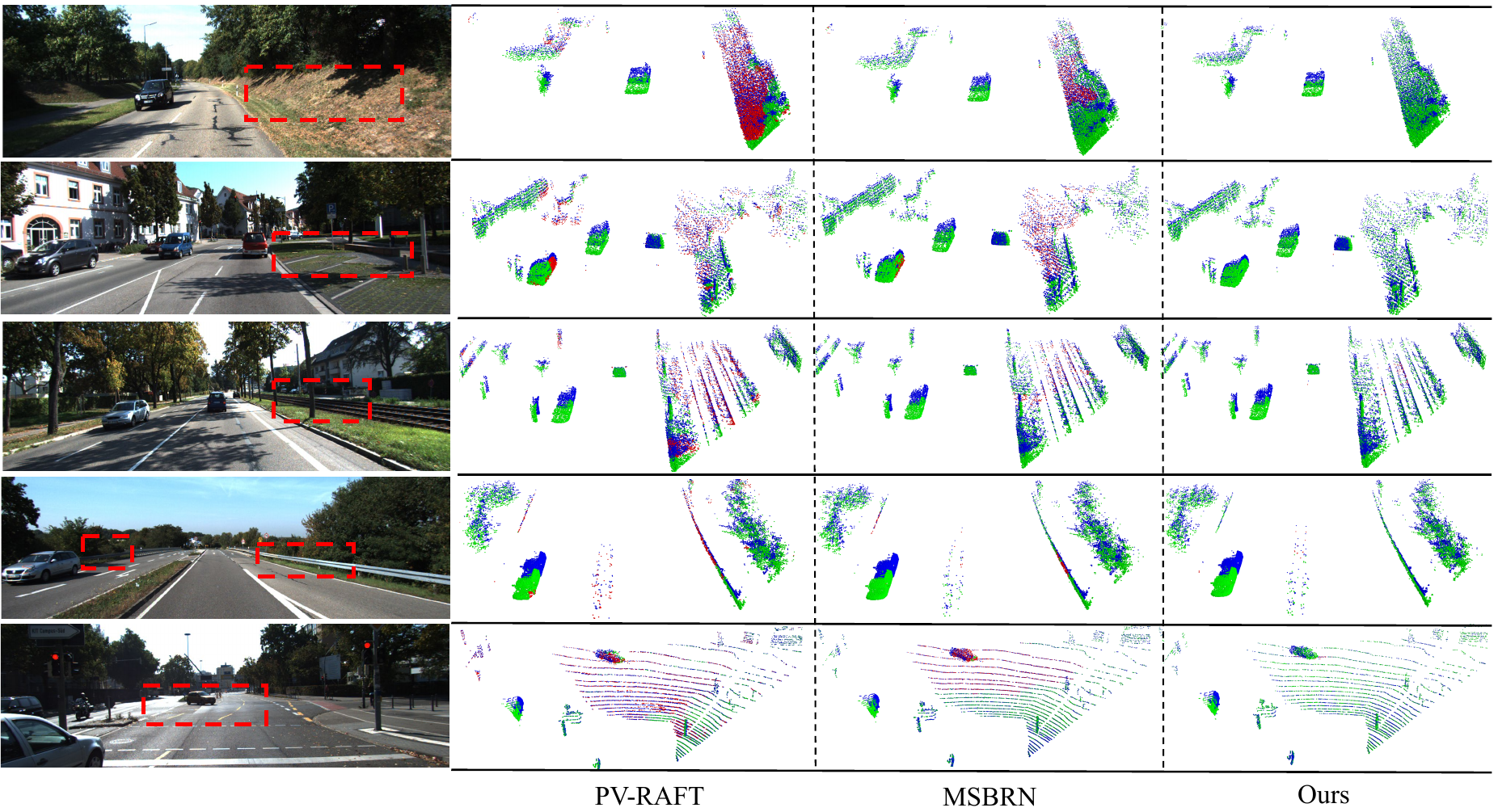}
\vspace{-8pt}
\caption{\textbf{Qualitative results on the test set of KITTI.} It shows that providing global motion propagation improves performance in areas with ambiguous geometric characteristics, such as embankments, roadside grassy areas, and some slender structures (curbs or tracks). \textcolor{blue}{Blue}, \textcolor{green!50!gray}{green} and \textcolor{red}{red} points respectively indicate \textcolor{blue}{the first frame $P_t$}, \textcolor{green!50!gray} {accurately estimated in $P_t$} and \textcolor{red}{inaccurately estimated in $P_t$ } (measured by Acc3DS).}

\vspace{-3pt}
\label{fig:visual1}
\end{figure*}

\noindent\textbf{Implementation Details.} All experiments were conducted using PyTorch~\cite{paszke2019pytorch}. For a fair comparison, we randomly sampled 8192 points as the network input. Consistent with prior work~\cite{cheng2023multi}, the point numbers at each layer are $N^1 = 2048$, $N^2 = 512$, $N^3 = 256$, and $N^4 = 64$. During training, we performed 4 update iterations per layer, which were also maintained during evaluation. We used the AdamW~\cite{kingma2014adam,loshchilov2017decoupled} optimizer with parameters $ \beta_1 = 0.9$ and $\beta_2=0.999$. The learning rate was adjusted using the CosineLR strategy, starting at 1e-3. The model was trained for a total of 300 epochs, and the same evaluation metrics as in recent studies~\cite{wu2020pointpwc,cheng2023multi,liu2024difflow3d} were employed.

\setlength{\tabcolsep}{2.5mm}
\begin{table*}[!t]
	\begin{center}
         \footnotesize
		
		\resizebox{1.0\textwidth}{!}        
            {
		\begin{tabular}{l|lccc|lccc}
		    \hline
			\toprule
			
			 &&\multicolumn{3}{c|}{\begin{tabular}[c]{@{}c@{}}FT3D$_{s}$\end{tabular}} &\multicolumn{3}{c}{\begin{tabular}[c]{@{}c@{}}KITTI$_{s}$\end{tabular}}\\ \cline{2-9}
			  \multirow{-2}{*}{\begin{tabular}[c]{@{}c@{}}Method \end{tabular}} & EPE3D\textcolor{red}{$\downarrow$} & Acc3DS\textcolor{green!60!gray}{$\uparrow$}& Acc3DR\textcolor{green!60!gray}{$\uparrow$}& Outliers\textcolor{red}{$\downarrow$}& EPE3D\textcolor{red}{$\downarrow$} & Acc3DS\textcolor{green!60!gray}{$\uparrow$}& Acc3DR\textcolor{green!60!gray}{$\uparrow$}& Outliers\textcolor{red}{$\downarrow$}\\ \midrule

		    PV-RAFT       & 0.0461   &  0.8169&0.9574   & 0.2924    &0.0560  & 0.8226  & 0.9372  &  0.2163  \\ 
		    
		      \cellcolor{gray!20}PV-RAFT(w/ISU)  
				&\cellcolor{gray!20}{\bf0.0331 \textcolor{red!50!gray}{($\downarrow$  28.2\%)}}  & \cellcolor{gray!20}\bf 0.9198
				&\cellcolor{gray!20}\bf 0.9783 & \cellcolor{gray!20}\bf 0.1802	
				 &\cellcolor{gray!20}{\bf0.0514 \textcolor{red!50!gray}{($\downarrow$ 8.2\%)}} & \cellcolor{gray!20}\bf 0.8952
				&\cellcolor{gray!20}\bf 0.9576 & \cellcolor{gray!20}\bf 0.1672	
				\\

			PT-FlowNet     & 0.0304 & 0.9142 & 0.9814   &   0.1735    & 0.0224 &  0.9551  &0.9838    &    0.1186   
            \\ 
             \cellcolor{gray!20}PT-FlowNet(w/ISU)  
				&\cellcolor{gray!20}{\bf0.0242 \textcolor{red!50!gray}{($\downarrow$20.4\%)}} &\cellcolor{gray!20}\bf0.9494 
				&\cellcolor{gray!20}\bf0.9860&\cellcolor{gray!20}\bf0.1166
				& \cellcolor{gray!20}{\bf0.0209  \textcolor{red!50!gray}{($\downarrow$6.7\%)}}& \cellcolor{gray!20}\bf0.9553
				&\cellcolor{gray!20}\bf 0.9846 &  \cellcolor{gray!20}\bf0.1137
				\\

            MSBRN     & 0.0150 & 0.9730 & 0.9920   &   0.0243  & 0.0110 & 0.9710 & 0.9890   &   0.0850   
            \\
             
            \cellcolor{gray!20}MSBRN(w/ISU)   & \cellcolor{gray!20}{\bf0.0107 \textcolor{red!50!gray}{($\downarrow$28.6\%)}} & \cellcolor{gray!20}\bf0.9741   &   \cellcolor{gray!20}\bf0.9927   &  \cellcolor{gray!20}\bf0.0892 & \cellcolor{gray!20}{\bf0.0104 \textcolor{red!50!gray}{($\downarrow$5.4\%)}} & \cellcolor{gray!20}\bf0.9718& 
             \cellcolor{gray!20}\bf0.9895  &  
             \cellcolor{gray!20}\bf0.0829   
            \\

			\bottomrule
			\hline
		\end{tabular}
		}
	\vspace{-6pt}
	\caption{\textbf{The plug-and-play capability of our methods.} Our Iterative SSM-based Update Module (ISU) can effectively improve the accuracy introduced into recent methods on both FlyingThings3D and KITTI datasets. The best results are in \bf{bold}. }
		
		\label{table:plug}
	\end{center}
	\vspace{-8mm}
\end{table*}

\subsection{Comparisons with State-of-the-art}

We validate the superiority of FlowMamba on datasets with two distinct pre-processing conditions, one including occlusion \cite{liu2019flownet3d} and the other excluding it \cite{gu2019hplflownet}, respectively.

\noindent\textbf{Comparison on point clouds without occlusion.} We compare our FlowMamba with a series of State-of-the-art (SOTA) methods on the FlyingThings3D and KITTI without occlusion. Table \ref{table:nocc} showcases the superiority of our FlowMamba approach, surpassing all previous methods across both 3D and 2D evaluation metrics. In comparison to the state-of-the-art DifFlow3D method~\cite{liu2024difflow3d}, our model exhibits a notable reduction in EPE3D by 21.9\% on the FlyingThings3D and 20.5\% on the KITTI, respectively. FlowMamba also has excellent generalization capabilities that achieve a millimeter-level EPE3D of 0.0062m on the KITTI dataset, just training on the synthetic FlyingThings3D dataset. Notably, we have achieved millimeter-level precision on both datasets for the first time.

\noindent\textbf{Comparison on point clouds with occlusion.} We also evaluate FlowMamba on the datasets with occlusion and compare it with the SOTA results. Experiments in Table \ref{table:occ} demonstrate that FlowMamba outperforms all previous methods on the FlyingThings3D and KITTI. In comparison to the state-of-the-art DifFlow3D method~\cite{liu2024difflow3d}, our model achieves a remarkable reduction in End-Point Error (EPE3D) by 20.9\% and 9.6\% on the FlyingThings3D and KITTI datasets, respectively.

\setlength{\tabcolsep}{1.5mm}
\begin{table}[!t]
	\footnotesize
	\begin{center}
		\resizebox{1.0\columnwidth}{!}
		{
	\begin{tabular}{l||cccc}
	            \hline
				\toprule
				Method& EPE3D\textcolor{red}{$\downarrow$} & Acc3DS\textcolor{green}{$\uparrow$}& Acc3DR\textcolor{green}{$\uparrow$}& Outliers\textcolor{red}{$\downarrow$}\\
				\noalign{\smallskip}
				\hline\hline
				\noalign{\smallskip}

				Conv GRU    
				&0.0134 &0.9763
				&0.9931 & 0.0466\\

				Mamba    
				&0.0141 & 0.9751
				&0.9929 & 0.0482\\

                    Bi-Mamba    
				&0.0127 & 0.9799
				&0.9935 & 0.0353\\

                    ISU(Ours)    
				&0.0101 & 0.9835
				&0.9946 & 0.0286\\

				ISU + FIO (Ours)
				 &\bf0.0089 & \bf 0.9861
				&\bf 0.9955 & \bf 0.0243
				\\
				\cline{1-5}\noalign{\smallskip}

				\bottomrule
				\hline
            	
			\end{tabular}
		}
	
	\vspace{-8pt}
	\caption{\textbf{Ablation studies the effectiveness of proposed modules on FlyingThings3D.}}
	\vspace{-3mm}
	\label{table:ablation of all}
	\end{center}
\end{table}

\subsection{Universality of proposed modules.}
\label{sec:plug}

Our proposed ISU can be used as a plug-and-play module to improve the accuracy of several iterative-based methods~\cite{wei2021pv,fu2023pt,cheng2023multi}. As shown in Table \ref{table:plug}, we replace their GRU with our ISU, resulting in significant improvements in baseline accuracy on both the FlyingThings3D and KITTI datasets. After incorporating ISU, the EPE3D of PV-RAFT decreased by $28.2\%$ on FlyingThings3D and $8.2\%$ on KITTI, respectively. 
The EPE3D of PT-FlowNet was reduced by $20.4\%$ and $6.7\%$, respectively.
Although MSBRN has achieved impressive results, our ISU has demonstrated the potential to further boost its performance significantly ($28.6\%$ and $5.4\%$ respectively).

\setlength{\tabcolsep}{0.2mm}
\begin{table}[!t]
	\footnotesize
	\begin{center}
		\resizebox{1.0\columnwidth}{!}
		{
	\begin{tabular}{l||cccc}
	            \hline
				\toprule
				Component& EPE3D\textcolor{red}{$\downarrow$} & Acc3DS\textcolor{green}{$\uparrow$}& Acc3DR\textcolor{green}{$\uparrow$}& Outliers\textcolor{red}{$\downarrow$}\\
				\noalign{\smallskip}
				\hline\hline
				\noalign{\smallskip}

				w/o context feature
				&0.0096 & 0.9849 
				&0.9951 & 0.0265\\

				w/o correlation feature
				&0.0095 & 0.9851
				&0.9953 & 0.0264\\

                    w/o hidden information
				&0.0093 & 0.9853
				&0.9953 & 0.0259\\

				Ours (full, with all features)     
				 &\bf0.0089 & \bf 0.9861
				&\bf 0.9955 & \bf 0.0243\\
				\bottomrule
				\hline
            	
			\end{tabular}
		}
	
	\vspace{-8pt}
	\caption{\textbf{Ablation experiment of FIO strategy on FlyingThings3D.}}
	\vspace{-3mm}
	\label{table:ablation of FIO}
	\end{center}
\end{table}

\setlength{\tabcolsep}{2mm}
\begin{table}[!t]
\footnotesize
	\begin{center}
\resizebox{1\columnwidth}{!}
{  
\begin{tabular}{llcccc}  
\hline  
\multirow{2}{*}{Dataset}       & \multirow{2}{*}{Method} & \multicolumn{4}{c}{Number of Iterations}              \\ \cline{3-6}   
                               &                         & 1      & 2      & 3      & 4      \\ \hline  
\multirow{2}{*}{FT3D} & MSBRN                   & 0.0403 & 0.0241 & 0.0177 & 0.0152 \\  
                               & Ours                    & \bf0.0176 & \bf0.0118 & \bf0.0097 & \bf0.0089 \\ \hline  
\multirow{2}{*}{KITTI}         & MSBRN                   & 0.0452 & 0.0239 & 0.0143 & 0.0110 \\  
                               & Ours                    & \bf0.0204 & \bf0.0110 & \bf0.0070 & \bf0.0062 \\ \hline  
\end{tabular}  
}  
\vspace{-8pt}
	\caption{\textbf{Ablation study of the number of iterations.}}
	\vspace{-5mm}
	\label{table:iter}
	\end{center}
\end{table}

\subsection{Ablation Study}

In this section, we evaluate our model in different settings to verify our proposed modules in several aspects. All results are obtained using the same number of iterations as reporting in \cite{cheng2023multi}. Figure \ref{fig:visual1} illustrates the superiority of our proposed method. It demonstrates that our method improves performance in areas with ambiguous geometric characteristics, such as flat regions and slender structures (curbs or tracks).

\noindent\textbf{Effectiveness of proposed modules.} We compare different global information propagation settings to demonstrate the effectiveness of our design, as shown in Table \ref{table:ablation of all}. We first try using a standard unidirectional Mamba instead of the Conv GRU and find it brings a decline in performance, as the limitations of unidirectional sequence modeling in effectively capturing global dependencies. Subsequently, we replaced it with bidirectional Mamba, utilizing the bidirectional mechanism to ensure comprehensive aggregation of information from other points. However, due to the impact of global noise propagation, the performance improvement remains marginal. We then replaced it with the ISU, which can adaptively perform global information fusion and updating, significantly improving estimation accuracy. Additionally, we incorporated the FIO strategy, which constructs coherent spatial dependencies, further enhancing the global information propagation capabilities of the ISU.

\noindent\textbf{Feature-induced ordering.}
To verify the impact of ordering settings with different features on final performance, we compared the estimation results by removing various features. As shown in Table \ref{table:ablation of FIO}, the feature-induced ordering strategy enhances motion estimation accuracy by sorting the point cloud. Additionally, using all three features for estimation leads to more convincing results.

\setlength{\tabcolsep}{3mm}
\begin{table}[!t]
	\footnotesize
	\begin{center}
		\resizebox{0.6\columnwidth}{!}
		{
	\begin{tabular}{l||cc}
	            \hline
				\toprule
				Method & Parameters & Runtime\\
				\noalign{\smallskip}
				\hline\hline
				\noalign{\smallskip}
			
				
				Baseline  
				&6.1M
				&283ms \\
                    \cline{1-3}\noalign{\smallskip}
				Ours  
				&6.0M
				&304ms \\
                    \bottomrule
				\hline
            	
			\end{tabular}
		}
	
	\vspace{-8pt}
	\caption{\textbf{Runtime comparison.} The runtime comparison between the baseline and our FlowMamba with 4 iterations setting. The baseline, based on MSBRN, includes an additional encoder for extracting contextual features. }
	\vspace{-5mm}
	\label{table:time}
	\end{center}
\end{table}

\noindent\textbf{Number of iterations.}
Our FlowMamba can achieve better performance with a smaller number of iterations. As shown in Table \ref{table:iter}, our network gets the same performance in just 2 iterations compared to MSBRN with 4 iterations. This indicates that our module can filter and reduce noise in hidden states through global motion propagation.

\subsection{Parameters, and Runtime.}

We compare FlowMamba with baseline methods to show the efficiency of ISU. All evaluations were conducted on a single RTX 3090 GPU. As shown in Table \ref{table:time}
, FlowMamba demonstrates highly competitive efficiency. Our baseline, based on MSBRN \cite{cheng2023multi}, incorporates an additional encoder to obtain contextual features. Compared to the baseline, FlowMamba has 
fewer parameters, with only a slight increase in computational time, while achieving significant performance improvements as reported in Table \ref{table:nocc}.

\section{Conclusion}
\label{sec:conclusion}

We proposed a novel scene flow estimation method, called FlowMamba. Its core component, the ISU module, is designed to efficiently propagate matching information globally and model long-range motion dependencies. To alleviate the impact of point cloud irregularity and enhance the global propagation capabilities of the ISU, we introduced an FIO strategy to order points into a sequence with spatial continuity at a high level. Extensive experiments demonstrate the superiority of our FlowMamba and its generalization ability on FlyingThings3D and KITTI. It shows the critical importance of global information propagation for point cloud motion estimation. The proposed method also showcases a strong university as a plug-and-play module for various approaches.

\clearpage
\section{Acknowledgments}
This work is supported by the National Natural Science Foundation of China (62122029, 62472184), the Fundamental Research Funds for the Central Universities, and the National Natural Science Foundation of China (623B2036).

\bibliography{aaai25}

\end{document}